\documentclass[conference]{IEEEtran}
\IEEEoverridecommandlockouts
\usepackage{cite}
\usepackage{url}
\usepackage{comment}
\usepackage{amsmath,amssymb,amsfonts}
\usepackage{algorithmic}
\usepackage{graphicx}
\usepackage{textcomp}
\usepackage{xcolor}
\usepackage{hyperref}

\def\BibTeX{{\rm B\kern-.05em{\sc i\kern-.025em b}\kern-.08em
    T\kern-.1667em\lower.7ex\hbox{E}\kern-.125emX}}
\begin{document}

\title{One-shot learning for the long term: consolidation with an artificial hippocampal algorithm}

\author{
\IEEEauthorblockN{Gideon~Kowadlo}
\IEEEauthorblockA{\textit{Cerenaut}\\
Melbourne, Australia \\
gideon@cerenaut.ai}
\and
\IEEEauthorblockN{Abdelrahman Ahmed}
\IEEEauthorblockA{\textit{Cerenaut}\\
Sydney, Australia \\
abdel@cerenaut.ai}
\and
\IEEEauthorblockN{David Rawlinson}
\IEEEauthorblockA{\textit{Cerenaut}\\
Melbourne, Australia \\
dave@cerenaut.ai}}

\maketitle

\newcommand{\HF}{HF}

\begin{abstract}
Standard few-shot experiments involve learning to efficiently match previously unseen samples by class.
We claim that few-shot learning should be long term, assimilating knowledge for the future, without forgetting previous concepts.
In the mammalian brain, the hippocampus is understood to play a significant role in this process, by learning rapidly and consolidating knowledge to the neocortex incrementally over a short period. 
In this research we tested whether an artificial hippocampal algorithm (AHA), could be used with a conventional Machine Learning (ML) model that learns incrementally analogous to the neocortex, to achieve one-shot learning both short and long term.
The results demonstrated that with the addition of AHA, the system could learn in one-shot and consolidate the knowledge for the long term without catastrophic forgetting.
This study is one of the first examples of using a CLS model of hippocampus to consolidate memories, and it constitutes a step toward few-shot continual learning.
\end{abstract}

\begin{IEEEkeywords}
Few-shot learning, One-shot learning, CLS, Hippocampus, Computational model, Consolidation, Omniglot, AHA
\end{IEEEkeywords}

\section{Introduction}
\label{sec:introduction}
    Conventional ML models are trained on large i.i.d. datasets. Sample efficient few-shot learning remains challenging \cite{Lake2019} and additional training on new categories can cause catastrophic forgetting of existing representations \cite{Kumaran2016}. 
    Yet animals can learn new tasks from as few as one experience, with little effort. 
    
    The discrepancy between few-shot learning in machines and animals has led to a lot of interest in recent years, see \cite{Lake2019} for a survey.
    But standard few-shot learning experiments test the ability to match between a current set of samples, as opposed to assimilating knowledge of a new class permanently.
    We argue that these experiments do not go far enough.
    Few-shot learning should be learning for the long term, incorporating new knowledge to be used at a later time, without forgetting previously learnt categories.
    
    In the standard testing framework for few-shot classification, established in seminal papers \cite{Lake2015, Vinyals2016}, there is a period of pre-training where a feature extractor learns a set of base categories from a large number of samples.
    In the evaluation phase, one or few samples of a small number of novel categories are given as training examples.
    Next, test samples of the same categories are presented, and the task is to correctly identify the samples from train and test sets that belong to the same category.
 
    A standard model for learning in the mammalian brain is CLS (Complementary Learning System) \cite{Kumaran2016, OReilly2014}. In CLS, Neocortex and Hippocampal Formation (HF) comprise two complementary learning systems with bidirectional connections. Neocortex graudally learns structured representations of the environment and HF learns specifics quickly (one-shot). HF slowly consolidates memories into Neocortex without interference with existing memories (continual learning), through interleaved replay of stored memory patterns. Replay occurs when the animal is in a passive state, and it also occurs as a response to external cues when in an active state. In this model, the HF constitutes short-term memory (STM) and Neocortex constitutes long-term memory (LTM). 

    In previous work, we introduced an Artificial Hippocampal Algorithm (AHA) \cite{Kowadlo2019, Kowadlo2020} based on CLS, with extensions that allow it to generalise as well as separate, spatial input. 
    That work tested the Hippocampal Formation's ability to learn fast and temporarily, and did not extend to consolidating that knowledge back into the neocortex for long term knowledge. It was tested on the standard few-shot testing framework.
    
    In this study, we investigate whether it is possible to use AHA to complement a conventional, incrementally learned ML model (LTM) in a way that is analogous to the interaction of HF and Neocortex, to provide one-shot learning capability of new classes, for both short and long term.
    
    \subsection{Related Areas}
    Simulating hippocampal memory consolidation is also of interest from a cognitive computational neuroscience perspective. To the authors' best knowledge, the studies that model rapid learning of distinct memories do not explore consolidation into long term memory \cite{Norman2003,Ketz2013,Greene2013,Rolls2013,Schapiro2017a,Rolls1995}.
    Building models that replicate biological consolidation can test theory and illuminate unknown assumptions, improving our understanding.
    
    This research is also strongly related to continual learning, in which new tasks are introduced incrementally. The challenge is to learn the new tasks and avoid forgetting old tasks. See \cite{De2019,Maltoni2019} for surveys.
    Hippocampus inspired replay methods can be used. For example, a deep generative model (GAN) was used to mimic past data for interleaved training of the `task solving' model and applied to static images \cite{Shin2017}. In another example, a growing recurrent network constituting an episodic module, was used to learn instances that were replayed to a semantic module and applied to video input \cite{Parisi2018}.
    
    However, there are very few continual learning studies that test learning new categories from one or a few samples without forgetting. 
    A framework for continual few-shot learning was introduced recently in \cite{Antoniou2020}, and existing few-shot learning algorithms were applied to a range of tasks.
    As this paper investigates consolidation of new one-shot categories, it is a step toward continual few-shot learning.

\section{Related Work}
    \label{sec:related_work}
    The most common approach for solutions to few-shot learning, is meta-learning, implemented with neural networks such as Siamese networks \cite{Koch2015a}, matching networks \cite{Vinyals2016}, and prototypical networks \cite{Snell2017}, or Bayesian approaches \cite{Lake2011,Lake2015}. 
    In each case, weights are modified in an outer loop, to optimise performance on the few-shot task performed in the inner loop.
    An embedding and/or filter for comparing representations are learnt. 
    The representations can be interpreted as primitive features that are used to compose novel categories and thus compare samples by their component features.
    
    Other researchers have looked at learning novel few-shot classes without forgetting a base set of classes, sharing the objective of this paper \cite{Hariharan2017, Gidaris2018, Wang2018lowshot}. 
    One approach is to generate synthetic samples from the few, to augment training data, with representation regularization \cite{Hariharan2017} and combined with meta-learning approaches \cite{Wang2018lowshot}.
    In \cite{Gidaris2018}, a feature extractor and classifier is augmented with an attention based few-shot classification weight generator, utilising previously learnt features (on the base classes).
    In each of these examples, there is a feature extractor that learns from the base classes, and a classifier for the few-shot learning task. Classifier learning occurs with the base and few-shot classes together, so that they must all be available at the same time. Therefore, despite the shared goal of `not forgetting', it is less clear how these approaches will extend to continually adding new classes.

\section{Model}
\label{sec:model}
\begin{figure}
        \centering
        \includegraphics[width=0.6\columnwidth]{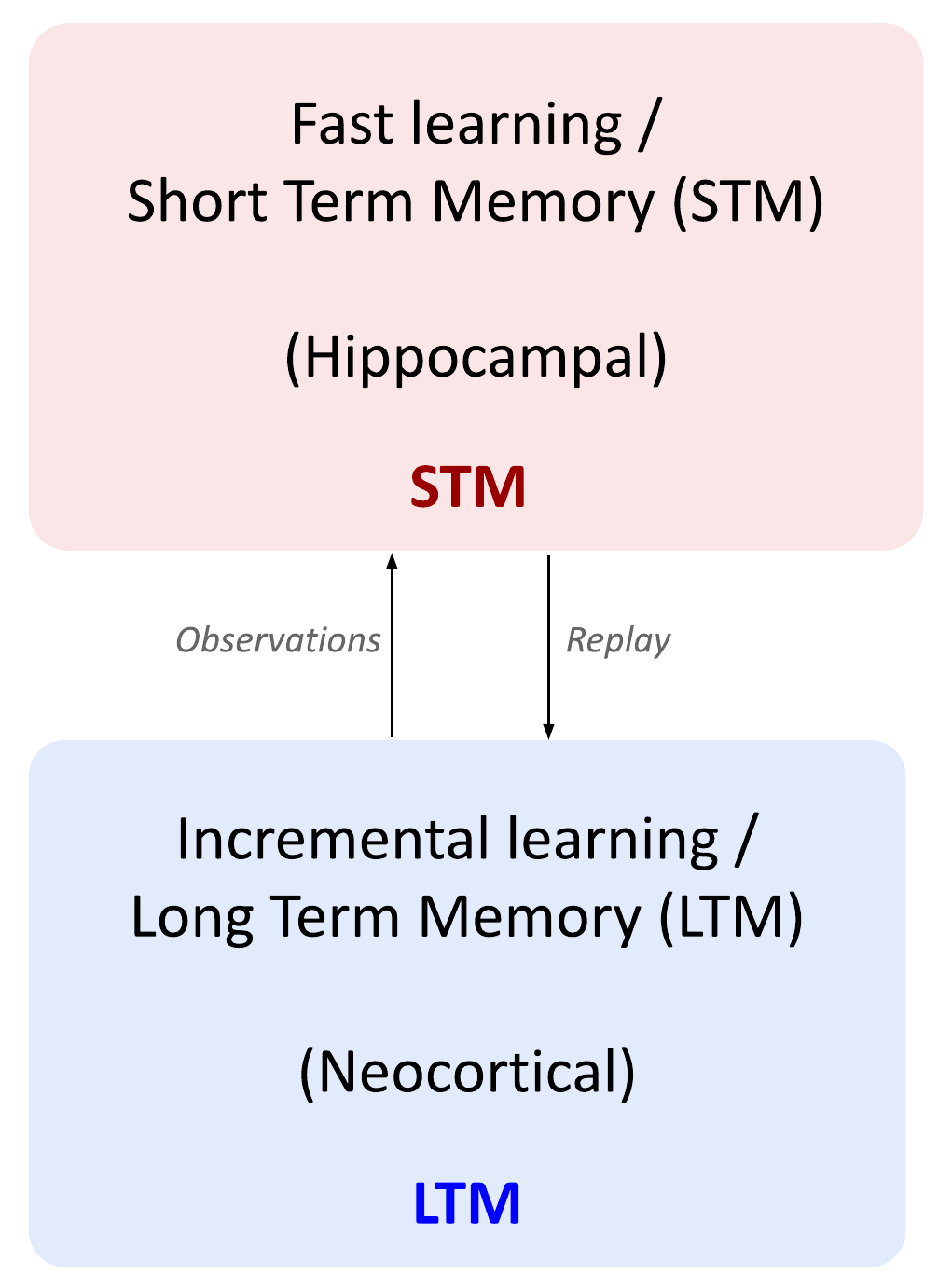}
        \caption{\textbf{System diagram:} A conventional Long Term Memory augmented with a complementary Short Term Memory module. Figure reproduced from \cite{Kowadlo2020}}
        \label{fig:system}
    \end{figure}

    The system consists of a simple ML model which acts as long term memory (LTM) and a complementary short term memory (STM) module, shown in Figure~\ref{fig:system}.
    Each component is described, followed by a description of how they work together.\footnote{The codebase is also publicly available on GitHub at: \url{https://github.com/Cerenaut/aha}}

    \subsection{LTM} 
    The LTM learns representations incrementally, with an unsupervised single layer sparse convolutional autoencoder (SCAE). A softmax layer is used as a classifier, trained with supervised learning to classify the autoencoder hidden layer according to image labels. See Appendix~\ref{app:scae} for details.

    \subsection{STM} The STM comprises AHA, an Artificial Hippocampal Algorithm, reported in detail in \cite{Kowadlo2019, Kowadlo2020}. 
    In contrast to the slow statistical learning of LTM, AHA learns rapidly from only one experience.
    AHA encodes samples with sparse representations that do not overlap or interfere with each other. This is achieved with both pattern separation and pattern retrieval pathways, providing the ability to separate highly similar inputs as well as generalize (see \cite{Kowadlo2020}). 
    
    AHA operates like an auto-associative memory. It uses an input cue to recall and `reconstruct' a corresponding memory. AHA itself learns without externally provided labels. However, in this work, AHA is modified to memorize and recall labels in addition to input image samples, referred to as (Image, Label) pairs. The (Image, Label) pairs are later used to train the LTM with supervised learning.
    
    More details are given in Appendix~\ref{app:aha}.

    \subsection{System} 
\begin{figure*}
        \centering
        \includegraphics[width=0.75\textwidth]{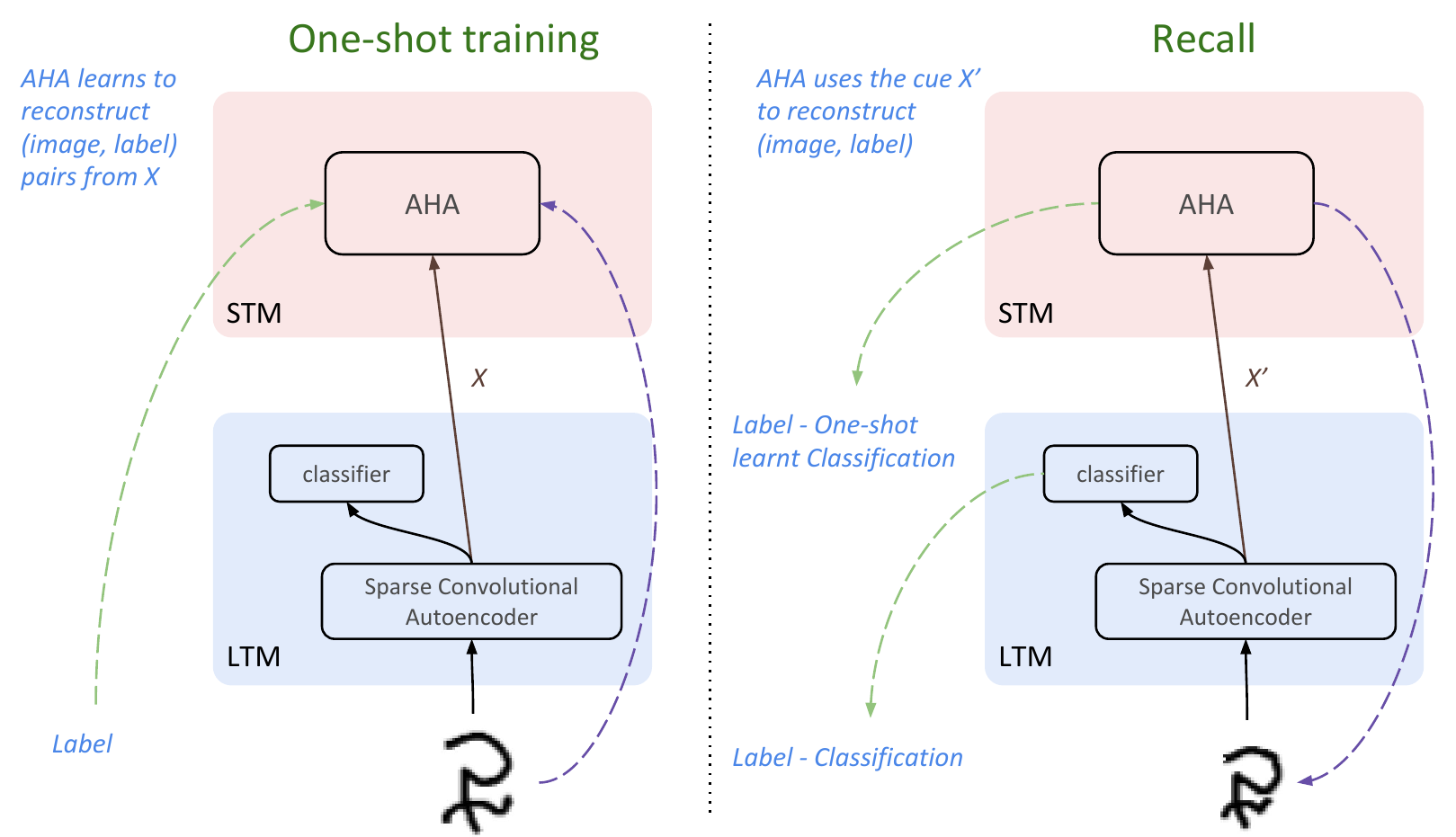}
        \caption{\textbf{AHA+LTM in operation after pre-training:} During training, AHA learns to reconstruct Label and Image in one-shot. While in STM, a different character-image cues recall of the learnt Image and Label and the labels are interpolated for `Short Term Inference'. Offline replay (Section~\ref{sec:consolidation}) consolidates memory into LTM, without causing forgetting of previous knowledge.}
        \label{fig:architecture}
    \end{figure*}

    Operation of the system is illustrated in Figure~\ref{fig:architecture}.
    The LTM is pre-trained to learn visual features common to the dataset. After pre-training, when presented with new samples (including novel, previously unseen categories), AHA receives LTM feature representations as input and learns combinations of these features in one-shot (i.e. from a single exposure). Subsequently, these categories will be recognised by AHA prompting replay to LTM.
    
    The replayed (Image, Label) pairs are used immediately for Short Term Inference, as well as for consolidating knowledge into LTM for Long Term Inference.
    
    \subsubsection{Short Term Inference}
    \label{sec:sti}
    Immediately after AHA has been shown a new (Image, Label) pair, it is memorized and available for recall and replay to LTM.
    While contained in the short term memory, it is available for Short Term Inference.
    For a given test image, the LTM provides a classification, and a label is retrieved from AHA (using the LTM autoencoder hidden activity as input to AHA).
    The classification probability distributions from both AHA and LTM's softmax classifier are interpolated to provide a classification.
    The final distribution consists of LTM and AHA both weighted 49.5\% and 1\% random uniform distribution.

    \subsubsection{Long Term Inference}
    \label{sec:lti}
    Knowledge is transferred from STM to LTM to enable Long Term Inference, a major challenge of this work.
    The LTM is a standard ML model that learns incrementally in a conventional training regime, sampling i.i.d. minibatches. 
    During one-shot learning, the system is only provided 1 new minibatch (containing a novel class). Adding new knowledge to the LTM from 1 sample is challenging as naively training on the same minibatch is expected to cause catastrophic forgetting. 

    The transfer occurs via an offline training phase referred to as Consolidation (explained in detail in Section~\ref{sec:consolidation}), where AHA is used as a type of generative model, independent of external input.
    In this phase, AHA is in a passive replay mode. Patterns of stored memories get replayed to LTM in a randomly interleaved fashion allowing minibatch training as if these were external inputs. 
    The memories are thereby \textbf{consolidated} to long term memory, LTM, and the short term memories can be overwritten. Thus LTM is able to learn new categories from only one exposure.

    \subsubsection{Consolidation}
    \label{sec:consolidation}
    The consolidation stage is achieved by repeatedly recalling random memories, `random recall' from AHA, populating a buffer, and then sampling from the buffer to replay the memories to LTM in training mode. 

    In `random recall', AHA recalls (Image, Label) pairs from internally generated random cues. The cues are generated by randomly sampling from a uniform distribution at each pixel location.
    The resulting outputs are fed back into the system to retrieve a crisper reconstruction and more accurate label classification (referred to as big-loop recurrence \cite{Kumaran2016}).
    A threshold is used to filter out remaining faint images.
    
    Due to the nature of retrieving samples from AHA using random cues, both Images and Labels are imperfect. There is noise, and sometimes a superposition of memories. 
    The remaining variation in Images and Labels may at first appear to be damaging, particularly as there are cases of incorrect labels. On the other hand, the variation may act as a regularizer and it is a form of data augmentation, which both assist in countering catastrophic forgetting.
    
    The `random recall' procedure is repeated for 25 steps per run, and the resulting (Image, Label) pairs are stored in a replay buffer for consolidation.
    The number of steps was chosen to make it highly probable to recall all of the memorized samples, and was optimised empirically.
    
    We randomly sample from the replay buffer (with a bias towards the unseen sample) and allow the LTM to train on these samples for 160 steps. 
    The replay buffer sampling mechanism ensures that there is at least one unseen sample (based on the label, as predicted by AHA).

\section{Experiments}

    We augmented a conventional LTM model with a hippocampal STM (Section~\ref{sec:model}) and tested it on one-shot supervised classification in two conditions a) Short Term Inference - with the help of memories stored in short term memory (STM), and b) Long Term Inference - after consolidation causes permanent weight changes to the LTM. A classification task was used to test the two forms of inference, and compared to a baseline case of LTM alone. Additionally, an upper bound on performance was measured with a second baseline of LTM, tested on classification \emph{without} any one-shot learning. 
    
\begin{figure}
        \centering
        \includegraphics[width=0.9\columnwidth]{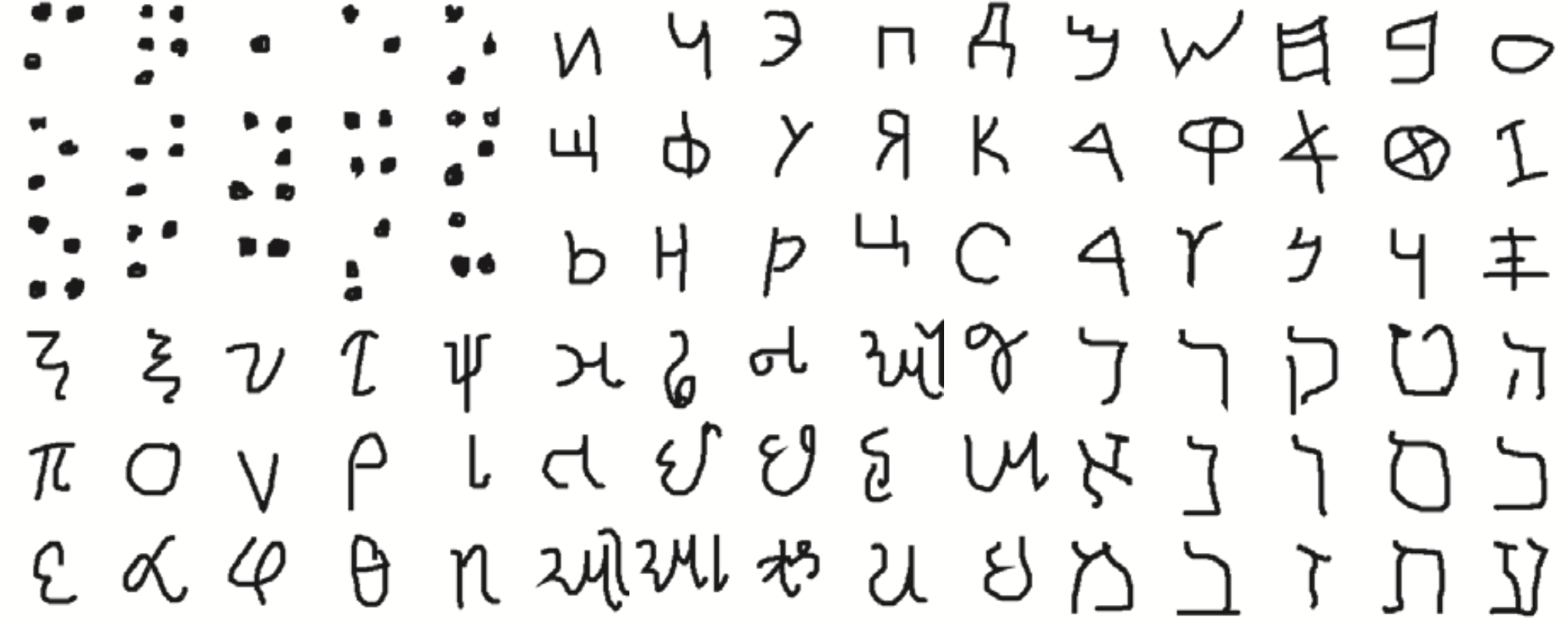}
        \caption{\textbf{Omniglot:} Handwritten characters from a range of alphabets of different styles. Modified from \cite{Lake2015}.}
        \label{fig:omniglot}
    \end{figure}

    The classification task was adapted from what has become a standard few-shot learning benchmark \cite{Lake2015}, using the Omniglot dataset. Omniglot is a collection of handwritten characters from 50 alphabets, a sample is shown in Figure~\ref{fig:omniglot}. 
    In the standard method, the system is pre-trained on a base set and then in an evaluation phase (the one-shot classification test itself), samples from a disjoint evaluation split are matched by category between a small train and test set (20 samples each). 
    
\begin{figure*}
        \centering
        \includegraphics[width=\textwidth]{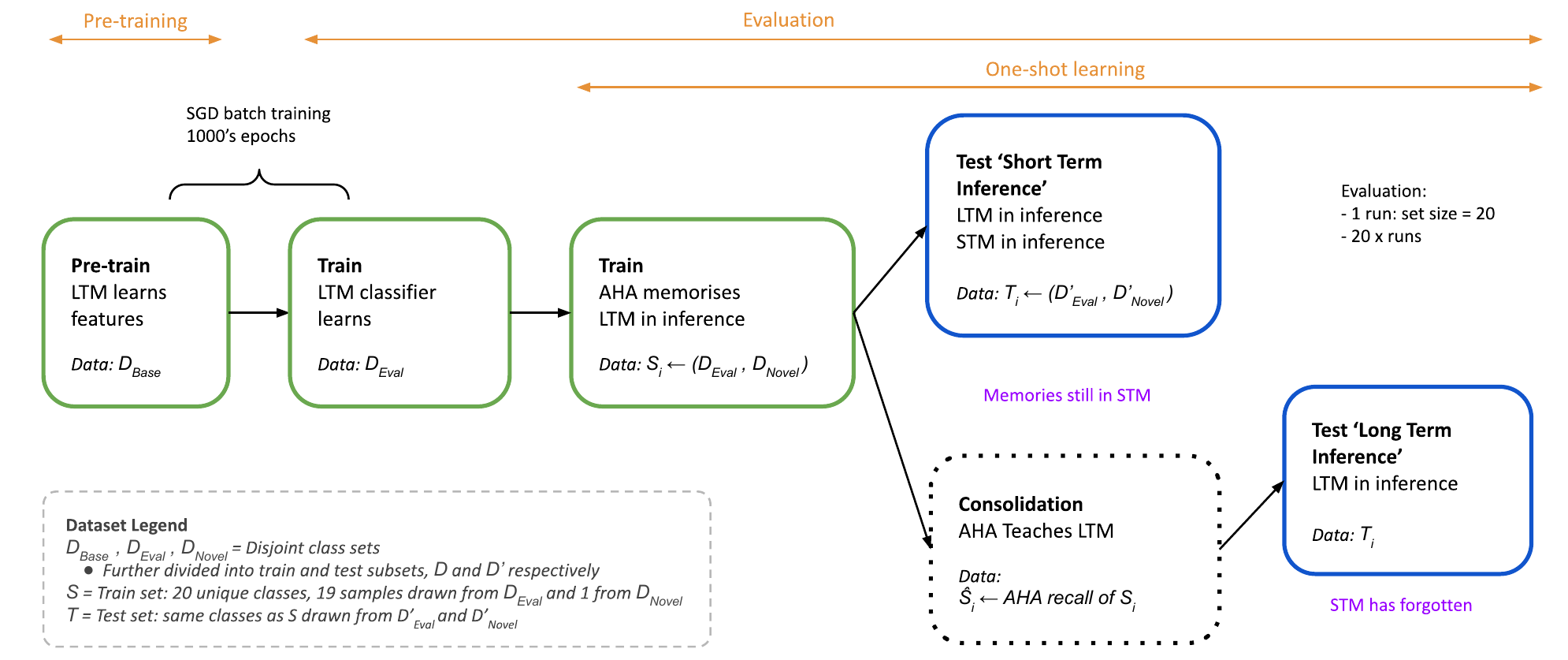}
        \caption{\textbf{Experimental Method Flowchart}}
        \label{fig:flow_method}
    \end{figure*}

    Our extended method is illustrated in a flowchart in Figure~\ref{fig:flow_method}.
    LTM is pre-trained with standard mini-batch training over many epochs.
    First, the LTM autoencoder learns with a pre-training set $D_{Base}$ where it learns the common features of the overall dataset.
    Then, the LTM classifier is trained on a disjoint set of classes, $D_{Eval}$.
    
    The one-shot learning evaluation phase begins with setting LTM to inference (weights are frozen) and training AHA on the features it receives from LTM, with support train sets $S_{i}$ of 20 samples. $S_{i}$ consist of 19 classes drawn from $D_{Eval}$ and 1 previously unseen class from disjoint set $D_{Novel}$. 
    We mix the novel category with known categories to replicate the way learning often occurs in embodied agents. 
    Although samples from $D_{Eval}$ are presented, there is only 1 for each class.
    Each $S_{i}$ is considered to be 1 `run'. 
    
    The system is tested before and after a consolidation phase, for Short Term Inference and Long Term Inference respectively.
    A target set with test samples of the same classes is presented to the network, and the LTM classifier provides predictions.
    See Model (Section~\ref{sec:model}) for details on the how the model performs these two forms of inference.
    
    The consolidation phase occurs without external input. The system has only been exposed to $S_{i}$ with 1 sample of each class. Consolidation must overcome the issue that the LTM learns incrementally over many i.i.d. minibatches, and is prone to forgetting with repeated exposure to a novel class. Consolidation is described in detail as part of the Model description, Section~\ref{sec:consolidation}.
    
    The evaluation phase (including pre-training of the LTM classifier) consists of 20 runs. Evaluation was repeated 10 times with different random seeds.

\section{Results}
    Results are summarised in Table~\ref{table:results}. 
    The `all classes' column includes 19 `previously learnt' classes (in pre-training) and 1 novel `one-shot' class (also referred to as `novel'), which was experienced only once. The `all-classes' score allows forgetting of existing classes to be measured in addition to learning of the novel class. `One-shot class' denotes accuracy on the novel category alone.
    Note that the impact of failing one-shot classification, a random guess at 1 out 19 classes, is $4.75\%$.
    Plus/minus refers to 1 standard deviation.
    
    Classification using the slowly learned representation of LTM achieves accuracy of $97.2\%$ overall, $45\%$ on the one-shot classes. Short Term Inference lowers accuracy to $81.40\%\pm{0.79\%}$ overall, but boosts it from $45\%$ to $74.50\%\pm{1.58\%}$ on one-shot classes. After consolidation the accuracy of LTM is increased to $97.47\%\pm{0.22\%}$ overall (back to the same accuracy as LTM alone) and $51.50\%\pm{4.12\%}$ on the one-shot classes.
    
    The LTM alone, trained and tested on all classes without the one-shot learning requirement, achieved $97.51\%\pm{0.23\%}$ accuracy, which is expected to be the upper bound on performance.
        
    
    \begin{table*}
    \caption{\textbf{Classification accuracy for baseline (LTM) and LTM+AHA:}. AHA dramatically improved one-shot learning performance and was able to consolidate that learning into Long Term Memory (LTM)}
    \begin{center}
    \def\arraystretch{1.5}%
    \begin{tabular}{p{0.2\textwidth} p{0.2\textwidth}  p{0.2\textwidth} p{0.2\textwidth}}
        \hline
        \textbf{Model/Test} & \textbf{Accuracy} \newline \textit{no one-shot learning} & \textbf{Accuracy} \newline \textit{all classes} &
        \textbf{Accuracy} \newline \textit{one-shot classes} \\ \hline
        LTM	& $97.51\%\pm{0.23\%}$ & $97.2\%$ & $45.0\%$ \\ \hline
        LTM+AHA \newline - Short Term Inference & NA & $81.40\%\pm{0.79\%}$ & $74.50\%\pm{1.58\%}$ \\ \hline
        LTM+AHA \newline - Long Term Inference & NA & $97.47\%\pm{0.22\%}$ & $51.50\%\pm{4.12\%}$ \\ \hline
    \end{tabular}
    \label{table:results}
    \end{center}
    \end{table*}

\subsection{Comparison to previous results}
The results are not directly comparable to the other approaches in the literature discussed in Related Work, Section~\ref{sec:related_work}.
They use a different paradigm, where a feature extractor is trained on base classes only, then a classifier is trained on base and novel (one-shot) classes. The classifier is not required to assimilate new knowledge at a later stage.
Additionally, they used a different dataset, Imagenet, enabled by much more complex feature extractors with orders of magnitude more parameters.

\section{Discussion}

As expected, the LTM on its own performs poorly at one-shot learning, although well above chance level, as it is designed to learn general statistical features incrementally. Nonetheless, it had high accuracy on `all-classes', close to the upper bound, because this only included 1 in 20 one-shot (novel) samples.

However, with the addition of AHA, it can perform one-shot learning on the supervised classification task, short and long term.

In the short term, accuracy for novel one-shot categories was convincingly improved, and was highest for one-shot classes over all conditions. 
Memory of novel classes are still held in AHA's memory, the LTM weights are not adapted and the system interpolates results between LTM and STM.
The boost occurs because relevant memories are effectively retrieved given appropriate cues i.e. subsequent unseen exemplar of the same class.
However, it comes at significant cost to the accuracy on existing knowledge, shown with the `all classes' accuracy.
AHA's recall is not perfect, and if it makes a mistake with either Image or Label recall, on a \textbf{non}-novel class, it will hurt performance. Overall, it evidently has a higher error rate than LTM classification, shown in the LTM `all-classes' condition. 
However, this is a trade-off determined by the method of interpolation.
Weighting interpolation toward STM would improve one-shot accuracy, and decrease `all classes', and vice versa. 
It would be possible to devise a measure of confidence to dynamically adjust the weighting - a possible area of future work.

With consolidation, accuracy on `one-shot classes' dropped, but showed a modest improvement on baseline accuracy. 
The `all classes' accuracy compared to lower accuracy on `one-shot classes' (i.e. it was not propped up this), shows that LTM did not forget existing knowledge, and in fact accuracy was restored to baseline performance, within the range of the expected upper bound.
Improvement on novel classes is modest, as consolidation from 1 experience is challenging. During consolidation, randomly sampled minibatches from the replay buffer have, on average, a very low proportion of novel classes compared to classes that are already represented. Given that there are errors in the recall process, the proportion is effectively lower than the 1 in 20 of the original experience.
On the other hand, this is also a likely explanation for the accuracy improvement on `all classes'. Encouragingly, the consolidation with interleaved replay (i.e. novel mixed with base classes) did prevent catastrophic forgetting.

There is large scope to improve consolidation, a topic of future work. For example, improving AHA itself, enforcing a higher proportion of novel classes in minibatches or modifying the learning rate. We expect to learn valuable principles from neuroscience, for example selectively biased replay emphasising samples that were most strongly encoded when first learnt \cite{POSKANZER2021107424} or via reward-based salience \cite{MichonF2019Hron}.

The use of a CLS architecture, in this case AHA, on one-shot supervised learning is novel. Previous work for AHA and other CLS studies involved the traditional `matching' framework (discussed in the Introduction, Section~\ref{sec:introduction}).
Additionally, the results demonstrate that hippocampal-inspired offline replay is a feasible approach to consolidate learning into long term memory (LTM). To the author's knowledge, this is a first for CLS models.

It is not surprising that the LTM, a conventional ML model, is not suited to one-shot learning.
However, it is an essential component of the overall one-shot learning system.
The LTM comprises a feature extractor, that learns about statistical generalities and it is complemented by an episodic memory that learns about specific memories from one experience.
The episodic `one-shot' memory, relies on the features that have been learnt over many samples.
This use of common features, or concepts, in new combinations for new experiences, is the basis of most one-shot/episodic learning, from hippocampus \cite{Ketz2013}, static classification \cite{Lake2015}, and Episodic RL \cite{botvinick2019}.
In this work, we also explored mapping abstracted episodes back to the feature space, and integrating them with existing representations, through additional statistical learning.

\section{Conclusion}

By augmenting a supervised classifier with a fast learning memory, it was possible to take one-shot learning beyond the transient `matching' paradigm discussed in the Introduction (Section~\ref{sec:introduction}), where classification consists of matching classes between a given train and test set, not persisting memories of one-shot classes.
A replay mechanism to consolidate knowledge through persistent weight adaptation, results in long term knowledge - in this case an ability to classify classes that have only been experienced once, without forgetting existing classes.
In addition, the results show how an artificial hippocampal algorithm can be used in a practical system to improve a standard supervised learning model, enabling it to learn new classes after only one exposure.
    
\section{Future Work}

Here we tested the ability to learn a single new category for the long term.
The next step is to test the system's ability to continually learn a stream of new categories (as opposed to 1 additional category), learnt in one-shot, to long term memory, without forgetting old categories.
We are currently using a similar replay mechanism in the context of a continual few-shot learning framework \cite{Antoniou2020} with both Omniglot and Slimagenet datasets.
These experiments are amongst the first to combine few-shot work with continual learning.

\bibliography{main.bib}
\bibliographystyle{IEEEtran}

\appendix

\subsection{LTM}
\label{app:scae}
The LTM is similar to the original architecture \cite{Kowadlo2019,Kowadlo2020}, with a sparse convolutional autoencoder (SCAE) and an interest filter. In this work, we added a supervised classifier. It is shown in the system diagram, Figure~\ref{fig:system_diagram_detail}.

The SCAE is based on the winner-take-all autoencoder from \cite{Makhzani2015} with two sparsity rules. First, a convolutional variant of the sparsity rule \cite{Makhzani2013}, where the top-k cells are chosen at each convolutional position by competition over all filters. And second, a lifetime sparsity rule to ensure that each filter is active at least once per batch.
We use a single autoencoder layer with tied weights. The layer was configured with 121 filters 10x10 receptive field, and a stride of 1. The layer is trained independently on the Omniglot background set to reconstruct the input using the mean-squared error (MSE) loss function. The weights are frozen when training the classifier and later AHA.

The interest filter, which mimics retinal processing, is used to suppress encoding of the Omniglot character's large amount of empty background; which is often represented strongly in the autoencoder hidden units.
The retina possesses centre-surround inhibitory and excitatory cells that can be well approximated with Difference of Gaussians (DoG) kernels \cite{Young1987}. Smoothing reduces sensitivity to feature locations.

The classifier is a single fully-connected layer with units equal to the number of classes. It is trained on the encoding of the previous layer using a softmax cross-entropy loss function for classification.

\subsection{STM}
\label{app:aha}

\begin{figure*}
\centering
\includegraphics[width=0.8\textwidth]{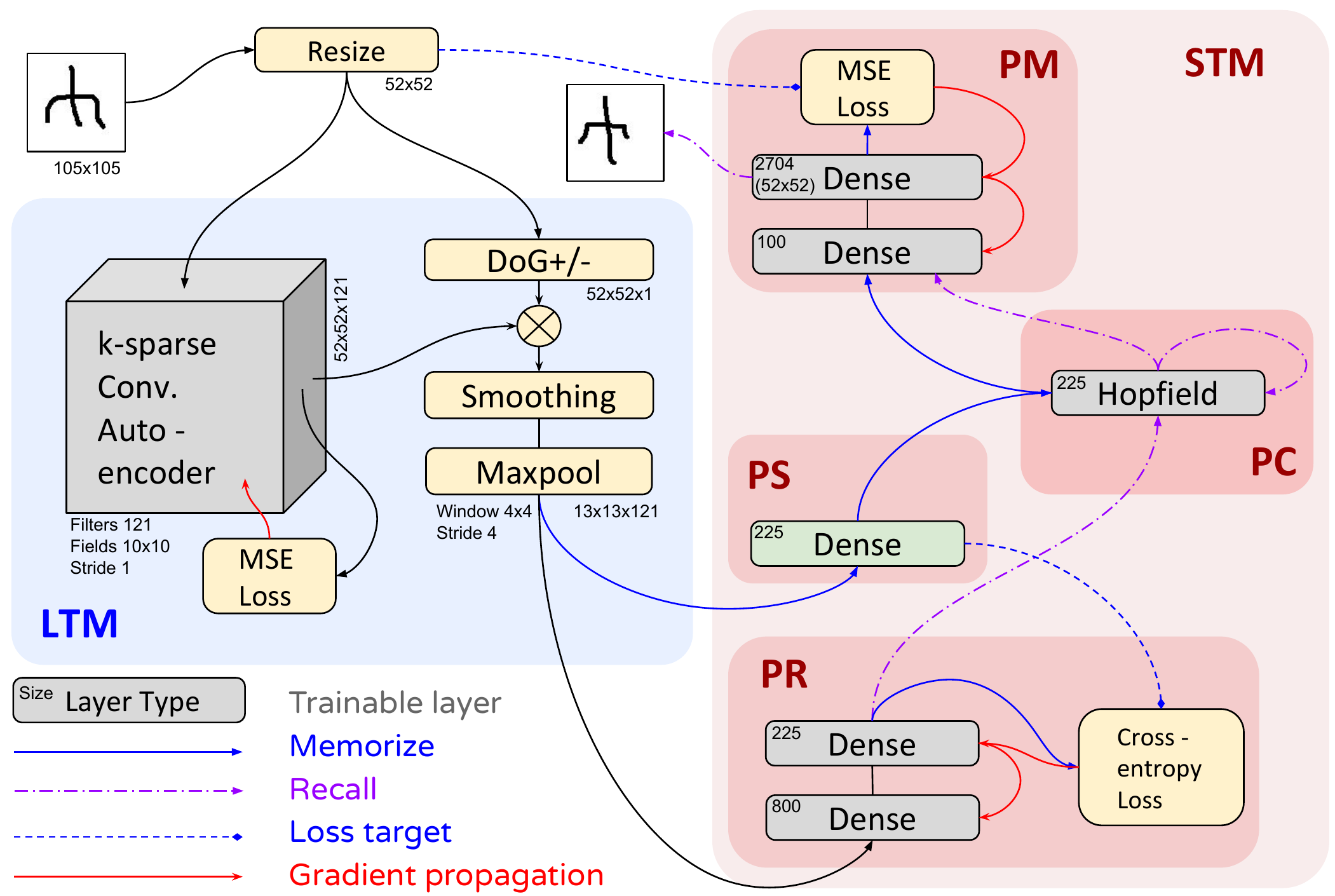}
\caption{\textbf{System diagram:} Local credit assignment via shallow backpropagation is used throughout. The dense layer in PS (green) is initialised, but not trained. Figure reproduced from \cite{Kowadlo2020}}
\label{fig:system_diagram_detail}
\end{figure*}

The version of AHA used, is similar to the one reported in \cite{Kowadlo2019,Kowadlo2020}. 
In this work we added an additional Pattern Mapper (PM) to reproduce labels, as well as the original image.
It is shown in the system diagram, Figure~\ref{fig:system_diagram_detail}, without the additional PM.

PS is the Pattern Separator. It creates non-interfering patterns from even highly overlapping similar observations. During training/encoding, those patterns are stored in PC, the Pattern Completer, an autoassociative memory module. At the same time, PR (Pattern Retriever) and PM (Pattern Mapper) are learning. PR learns to retrieve the patterns stored in PC, given the overlapping representations from LTM, whilst PM learns to do the inverse i.e. reproduce the original input from the PC pattern. At inference or `retrieval' time, a subsequent input causes PR to reproduce an appropriate cue for PC to retrieve a crisp version of the memorised `non-interfering' pattern. That unambiguous pattern is used by PM to reconstruct the original input.

PM and PR are 2 layer fully connected artificial neural networks. PC is a Hopfield network. PS is not trained, it is initialised with random weights and additionally implements temporal inhibition (mimicking neuron refractory periods) to ensure patterns do not overlap each other.
Each submodule PM, PC and PR is trained with local credit assignment only and no external labels are used for training. PM and PR are trained with self-supervised learning i.e. labels are internally generated.

Each of the submodules correspond to Hippocampus subfields or pathways.
PS = Dentate Gyrus (DG), PR = EC-CA3, PC = CA3 and PM = CA1.

\end{document}